\documentclass{article}

\RequirePackage{xcolor}
\usepackage{microtype}
\usepackage{graphicx}
\usepackage{subfigure}
\usepackage{booktabs} 

\usepackage{hyperref}

\usepackage[accepted]{icml2018}

\icmltitlerunning{Efficient Model--Based Deep Reinforcement Learning with Variational State Tabulation}

\usepackage{amsmath,amsfonts}
\usepackage[nolist]{acronym}
\usepackage[OT1]{fontenc}
\usepackage{verbatim}

\DeclareMathOperator*{\logit}{logit}

\begin{document} 

\twocolumn[
\icmltitle{Efficient Model--Based Deep Reinforcement Learning with\\ Variational State Tabulation}



\icmlsetsymbol{equal}{*}

\begin{icmlauthorlist}
\icmlauthor{Dane Corneil}{epfl}
\icmlauthor{Wulfram Gerstner}{epfl}
\icmlauthor{Johanni Brea}{epfl}
\end{icmlauthorlist}

\icmlaffiliation{epfl}{Laboratory of Computational Neuroscience (LCN),
School of Computer and Communication Sciences and Brain Mind Institute, School
of Life Sciences, \'{E}cole Polytechnique
    F\'{e}d\'{e}rale de Lausanne, Switzerland}

\icmlcorrespondingauthor{Dane Corneil}{dane.corneil@epfl.ch}

\icmlkeywords{Model-based, Reinforcement Learning, Deep networks}

\vskip 0.3in
]

\printAffiliationsAndNotice{}  

\begin{acronym}
\acro{RL}{Reinforcement Learning}
\acro{VaST}{Variational State Tabulation}
\acro{MFEC}{Model--Free Episodic Control}
\acro{NEC}{Neural Episodic Control}
\acro{MDP}{Markov Decision Process}
\acro{POMDP}{\emph{partially--observable} Markov Decision Process}
\acro{CNN}{Convolutional Neural Network}
\acro{DCNN}{Deconvolutional Neural Network}
\acro{VAE}{Variational Autoencoder}
\acro{LSH}{Locality Sensitive Hashing}
\end{acronym}

\newcommand{\dane}[1]{\textcolor{red}{Dane: #1}}
\newcommand{\johanni}[1]{\textcolor{red}{Johanni: #1}}

\begin{abstract}
    Modern reinforcement learning algorithms reach super--human performance on many board and video games, but they are sample inefficient, i.e. they typically require significantly more playing experience than humans to reach an equal performance level. To improve sample efficiency, an agent may build a model of the environment and use planning methods to update its policy. In this article we introduce \ac{VaST}, which maps an environment with a high--dimensional state space (e.g. the space of visual inputs) to an abstract tabular model. Prioritized sweeping with small backups, a highly efficient planning method, can then be used to update state--action values. We show how \ac{VaST} can rapidly learn to maximize reward in tasks like 3D navigation and efficiently adapt to sudden changes in rewards or transition probabilities.  
\end{abstract} 
\acresetall

\section{Introduction}

Classical \ac{RL} techniques generally assume a tabular representation of the state space~\cite{Sutton2018}. While methods like prioritized sweeping~\cite{Sutton2018,moore1993prioritized,peng1993efficient,van2013efficient} have proven to be very sample--efficient in tabular environments, there is no  canonical way to carry them over to very large (or continuous) state spaces, where the agent seldom or never encounters the same state more than once. Recent approaches to reinforcement learning have shown tremendous success by using deep neural networks as function approximators in such environments, allowing for generalization between similar states (e.g.~\citet{mnih2015human,mnih2016asynchronous}) and learning approximate dynamics to perform planning at decision time (e.g.~\citet{Silver2017,Oh2017,Farquhar2017,racaniere2017imagination,Nagabandi2017}). However, methods like prioritized sweeping that use a model for offline updates of $Q$-values (i.e. background planning \cite{Sutton2018}), have not yet been investigated in conjunction with function approximation by neural networks.

Adjusting the weights in a deep network is a slow procedure relative to learning in tabular environments. In particular, agents using deep architectures typically fail to take advantage of single experiences that significantly alter the policy. This was illustrated by recent work on \ac{MFEC} ~\cite{blundell2016episodic}, where a very simple agent using a semi--tabular approach significantly outperformed existing deep network approaches in the early stages of learning. The basic \ac{MFEC} agent uses a random projection from the observation space to a low--dimensional space, and stores the discounted returns associated with observations in a lookup table. The $Q$-values of states observed for the first time are determined by a k--Nearest--Neighbour average over discounted returns associated with similar existing states in the lookup table.

\begin{figure}[!ht]
\includegraphics[width=\columnwidth]{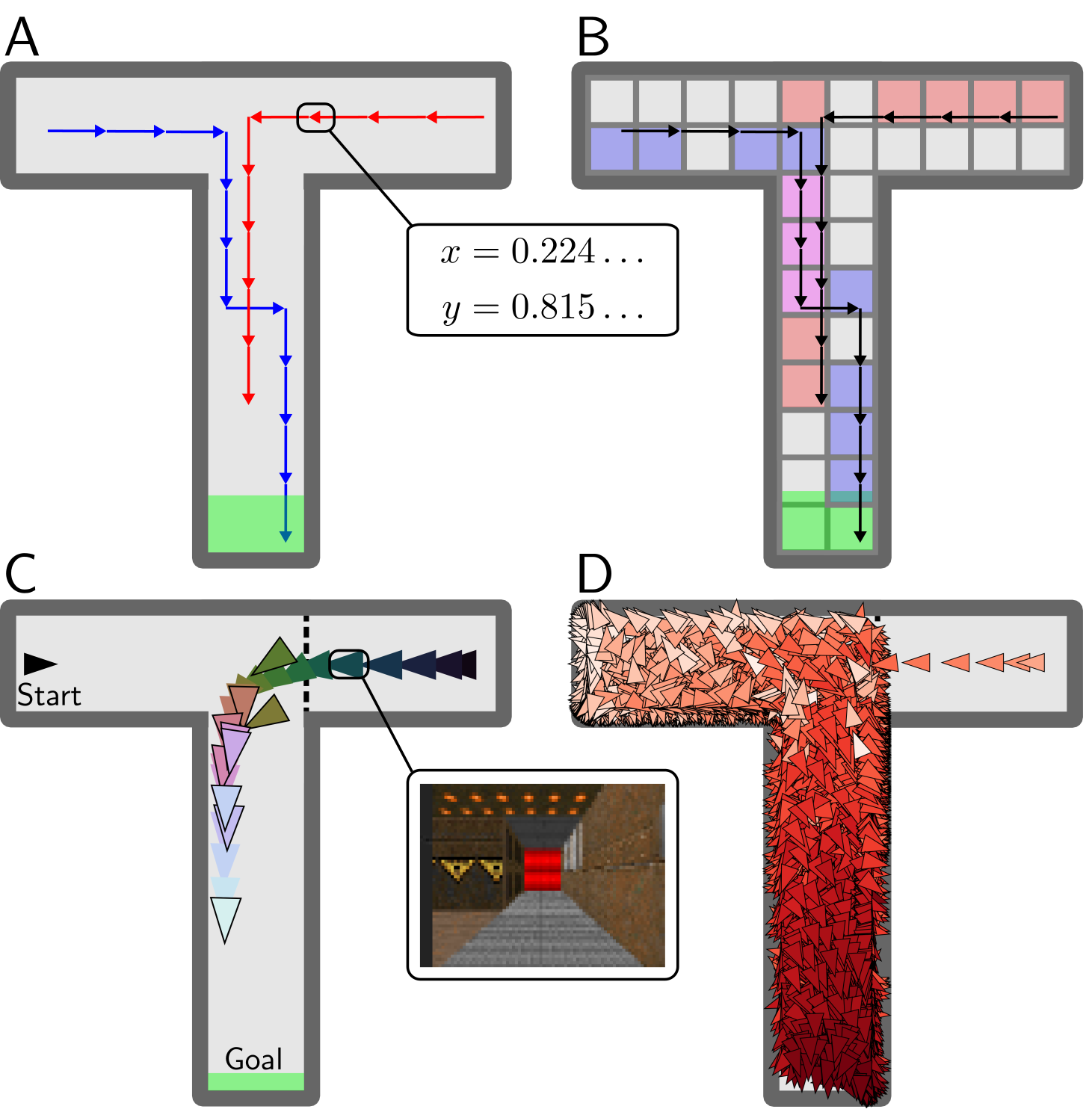}\vspace{-11pt}
\caption{\textbf{Using state tabulation for efficient planning}. [\emph{A}] Two episodes in a time discrete \ac{MDP} with a continuous state space, given by $(x,y)$ coordinates. [\emph{B}] The same episodes after discretising the state space by rounding. States visited on each trajectory are shaded; magenta states were shared by both trajectories, and can be leveraged by prioritized sweeping. [\emph{C}] 3D navigation with \ac{VaST}. The agent was trained to run from the start position to the goal, with one arm blocked (dotted line). After training, the agent experienced a trajectory from the blocked arm to the stem of the maze (arrows, no outline). If the observations in this trajectory mapped to existing states, the average coordinates and orientation where those states were previously observed are shown (matching colour arrows, outlined). 
The observations for one state are illustrated. [\emph{D}] A scatter plot of the values of all states after training according to the average position and orientation where they were observed; darker red corresponds to higher value. 
} 
\label{fig:introduction}
\end{figure}

As an example with an \ac{MFEC} agent, we can consider the T--maze task shown in \autoref{fig:introduction}A. The observations are continuous $(x,y)$ coordinates; the agent can take a fixed--sized step in one of four cardinal directions, with a rebound on hitting a wall, and a terminal reward zone (green). The first episode (red) is spontaneously terminated without reward; the discounted returns along the red trajectory are therefore set to zero. On the second episode (blue), the agent reaches the reward zone, and the discounted reward is immediately associated with the states visited along the blue trajectory. 

To improve sample efficiency, we consider how an agent could apply the
experience of the rewarded trajectory to $Q$-value estimates in the top--right arm. In particular, by learning a model of the environment, the agent could learn that both trajectories pass through the center of the T--maze, and that discovering a reward at the bottom of the maze should therefore change $Q$-value estimates in both of the arms at the top. This idea is exploited by the model--based \ac{RL} technique of prioritized sweeping~\cite{moore1993prioritized,peng1993efficient,van2013efficient}.

However, assuming random restarts, the agent in this task never encounters the same state more than once. In this case, given a deterministic task, prioritized sweeping as implemented by~\citet{van2013efficient} collapses to the \ac{MFEC} learning algorithm~\cite{brea2017}. 
We are therefore motivated to consider mapping the observation space to a
tabular representation by some form of discretisation. For example, with discretisation based on rounding the $(x, y)$ coordinates (a simple form of state aggregation \cite{li2006towards,Sutton2018}), the two trajectories in \autoref{fig:introduction}B now pass through several of the same states. A model--based prioritized sweeping algorithm would allow us to update the $Q$-values in the top--right arm of the maze to nonzero values after experiencing both episodes, despite the fact that the red trajectory did not result in reward.

If observations are given by high-dimensional visual inputs instead of $(x, y)$ coordinates, the simple form of state aggregation by rounding (\autoref{fig:introduction}B) is impractical. Instead, we propose and describe in this article the new method of \acf{VaST}\footnote{The full code for \ac{VaST} can be found at \url{https://github.com/danecor/VaST/}.} for learning discrete, tabular representations from high--dimensional and/or continuous observations. \ac{VaST} can be seen as an action conditional hybrid ANN--HMM (artificial neural network hidden Markov model, see e.g. \cite{Bengio1992,Tucker2017,Ng2016,Maddison2016}) with a $d$-dimensional binary representation of the latent variables, useful for generalization in \ac{RL}. \ac{VaST} is trained in an unsupervised fashion by maximizing the evidence lower bound. We exploit a parallelizable implementation of prioritized sweeping by small backups~\cite{van2013efficient} to constantly update the value landscape in response to new observations. By creating a tabular representation with a dense transition graph (i.e. where the same state is revisited multiple times), the agent can rapidly update state--action values in distant areas of the environment in response to single observations.



In \autoref{fig:introduction}C\&D, we show how \ac{VaST} can use the generalization of the tabular representation to learn from single experiences. We consider a 3D version of the example T--maze, implemented in the \emph{VizDoom} environment~\cite{Kempka2016ViZDoom}. Starting from the top--left arm, the agent was trained to run to a reward in the bottom of the T--maze stem. During training, the top--right arm of the T--maze was blocked by an invisible wall. After training, the agent observed a single, 20--step fixed trajectory (or ``forced run'')  beginning in the top--right arm and ending in the stem, without reaching the reward zone (\autoref{fig:introduction}C). The agent's early observations in the unexplored right arm were mapped to new states, while the observations after entering the stem were mapped to existing states (corresponding to observations at similar positions and orientations). The agent was able to update the values of the new states by prioritized sweeping from the values of familiar states (\autoref{fig:introduction}D), without needing to change the neural network parameters, as would be necessary with model-free deep reinforcement learners like DQN \cite{mnih2013playing}.

\section{Learning the Model}

In order to compute a policy using the model--based prioritized sweeping algorithm described by~\citet{van2013efficient}, we seek a posterior distribution $q(s_t|o_{t-k:t})$ over latent \emph{discrete} states $s_t$ given a causal filter over recent observations $o_{t-k:t}$. We use a variational approach to approximate this posterior distribution.


 

\subsection{The variational cost function}

For a sequence of states $s_{0:T} = (s_0, \ldots s_T)$ and observations $o_{0:T} = (o_0,\ldots o_{T})$, we consider a family of approximate posterior distributions $q_\phi(s_{0:T}|o_{0:T})$ with parameters $\phi$, which we assume to factorise given the current observation and a memory of the past $k$ observations, i.e. 
\begin{align}
    q_\phi(s_{0:T}|o_{0:T}) &= 
    \prod_{t=0}^Tq_\phi(s_t|o_{t-k:t})\, ,
\end{align}
where observations before $t=0$ consist of blank frames. To learn $q_\phi$, we also introduce an auxiliary distribution $p_\theta$ parameterized by $\theta$. Given a collection of $M$ observation sequences $\mathcal O = \{o^\mu_{0:T^\mu}\}_{\mu = 1}^M$ and hidden state sequences $\mathcal S = \{s^\mu_{0:T^\mu}\}_{\mu = 1}^M$, we maximize the log--likelihood $\log \mathcal L(\theta;\mathcal O) = \sum_{\mu=1}^M\log p_\theta\left(o^\mu_{0:T^\mu}\right)$ of the weight parameters $\theta$, while minimizing $\mathcal D_{KL}(q_\phi(\mathcal S|\mathcal O)||p_\theta(\mathcal S|\mathcal O))$. Together, these terms form the evidence lower bound (ELBO) or negative variational free energy 
\begin{align}
	\label{eq:negfree}
    \nonumber - \mathcal F(\theta, \phi; \mathcal O) &=
    \log \mathcal L(\theta;\mathcal O) - \mathcal D_{KL}(q_\phi(\mathcal S|\mathcal O)||p_\theta(\mathcal S|\mathcal O)) \\
    &= \mathbb{E}_{q_{\phi}}[\log p_\theta(\mathcal S,\mathcal O)] + \mathcal H(q_\phi(\mathcal S|\mathcal O))\, ,
\end{align}
where $\mathcal H$ denotes the entropy of the distribution. The term inside the expectation evaluates to 
\begin{align}
    \log p_\theta(\mathcal S,\mathcal O) =& 
    \sum_{\mu=1}^M \log \pi_{\theta_0}(s_0^\mu) 
    \nonumber + \sum_{\mu=1}^M \sum_{t=0}^{T^\mu} \log p_{\theta_{\mathcal{R}}}(o_t^\mu|s_t^\mu) \\
     &+ \sum_{\mu=1}^M \sum_{t=1}^{T^\mu} \log p_{\theta_\mathcal{T}}(s_t^\mu|a_t^\mu, s_{t-1}^\mu)\, ,
\end{align}
where $a_t^\mu$ denotes the action taken by the agent on step $t$ of sequence $\mu$, $\pi_{\theta_0}$ is the distribution over initial states, and $\theta_0 \cup \theta_\mathcal{R} \cup \theta_\mathcal{T} = \theta$.

We aim to learn the appropriate posterior distribution $q_\phi$ by minimizing the variational free energy (maximizing the ELBO). Our cost function from Eq.~\ref{eq:negfree} can be written as
\begin{equation}
    \mathcal F(\theta, \phi; \mathcal O) = \sum_{\mu = 1}^M \sum_{t=0}^{T^\mu}\big[\mathcal R^\mu_t + \mathcal T^\mu_t - \mathcal H^\mu_t\big],
\end{equation}
with reconstruction cost terms
\begin{equation}
    \mathcal R^\mu_t = - \sum_{s^\mu_t} q_\phi(s^\mu_t| o^\mu_{t-k:t})
    \log p_{\theta_{\mathcal{R}}}(o^\mu_t|s^\mu_t)\, ,
\end{equation}
transition cost terms
\begin{equation}
    \mathcal T^\mu_t = -\sum_{s^\mu_t, s^\mu_{t-1}}
    q_\phi(s^\mu_t, s^\mu_{t-1}| o^\mu_{t-k - 1:t})\log 
    p_{\theta_{\mathcal{T}}}(s^\mu_t| a^\mu_t, s^\mu_{t-1})
\end{equation}
for $t>0$ and $
    \mathcal T^\mu_0 = 
    -\sum_{s^\mu_0}q_\phi(s^\mu_0 | o^\mu_{0})\log 
    \pi_{\theta_0}(s^\mu_0)\, ,$
and entropy terms 
\begin{equation}
\mathcal H^\mu_t = 
-\sum_{s^\mu_t} q_\phi(s^\mu_t| o^\mu_{t-k:t})\log 
q_\phi(s^\mu_t| o^\mu_{t-k:t})\, .
\end{equation} 

\begin{figure}
\includegraphics[width=\columnwidth]{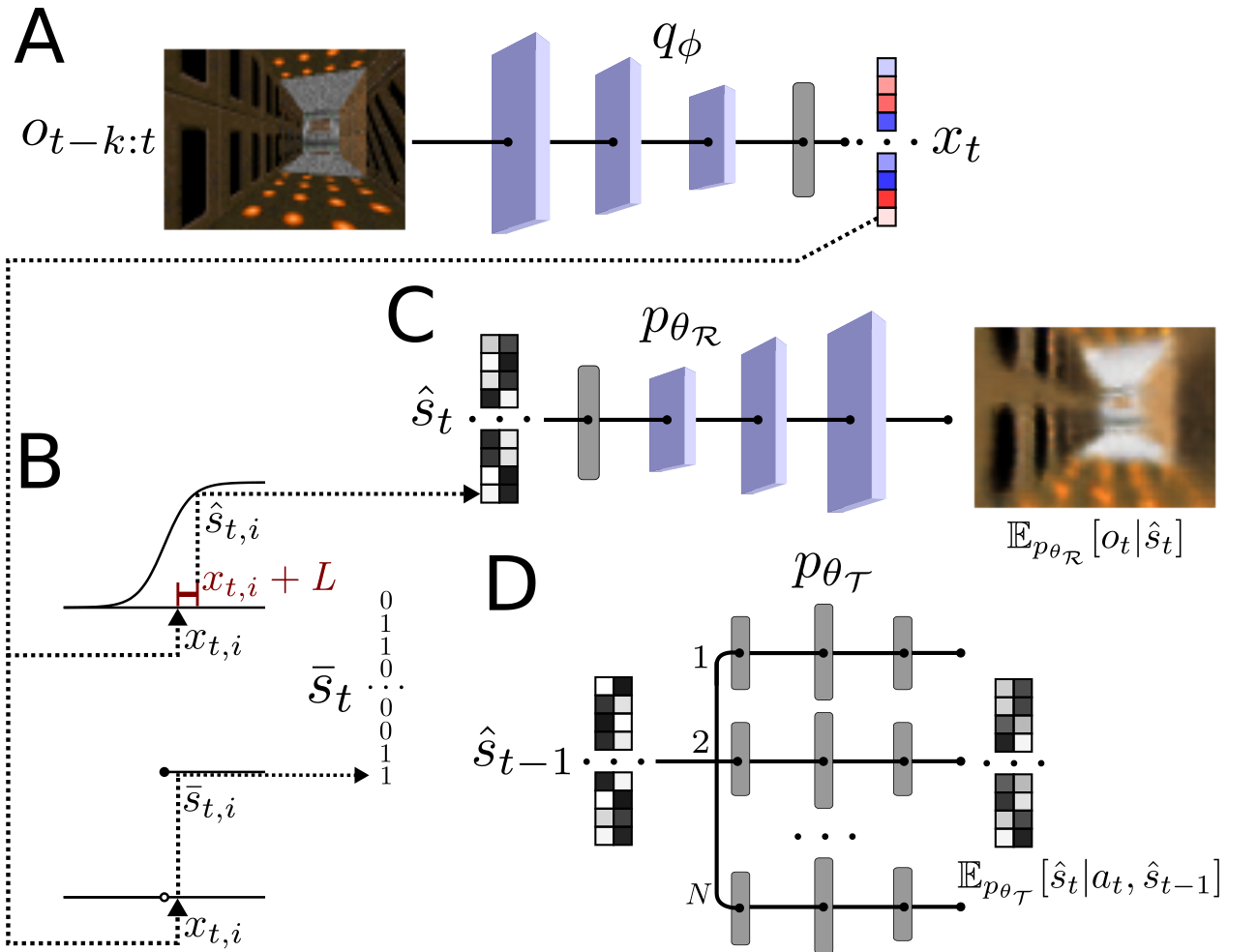}\vspace{-11pt}
\caption{\textbf{The network model}. 
    [\emph{A}] CNN encoder $q_\phi$. [\emph{B}] Encoder outputs can be used to sample each dimension from a Con--crete distribution for training ($\hat{s}_t$), or discretised to the Bernoulli mode $\bar{s}_t$ to update the table. The Con--crete distribution corresponds to a logistic activation with added noise $\mathit{L}$. [\emph{C}] DCNN decoder $p_{\theta_{\mathcal{R}}}$ and [\emph{D}] Transition network $p_{\theta_{\mathcal{T}}}$, with $N$ possible actions. For illustration, $\mathbb{E}_{p_{\theta_{\mathcal{R}}}}[o_t | \hat{s}_t]$ and $\mathbb{E}_{p_{\theta_{\mathcal{T}}}}[\hat{s}_t | a_t, \hat{s}_{t-1}]$ are shown.}
\label{fig:VaST}
\end{figure}

We parameterize the posterior distribution $q_\phi$ (or ``encoder'') using a deep \ac{CNN}~\cite{Krizhevsky2012imagenet}, and the observation model $p_{\theta_{\mathcal{R}}}$ using a deep \ac{DCNN}~\cite{goodfellow2014generative}, as shown in \autoref{fig:VaST}. We use a multilayer perceptron (3 layers for each possible action) for the transition model $p_{\theta_{\mathcal{T}}}$, and learned parameters $\theta_0$ for the initial state distribution $\pi_{\theta_0}$. The architecture is similar to that of a \ac{VAE}~\cite{kingma2013auto,rezende2014stochastic}, with the fixed priors replaced by learned transition probabilities conditioned on previous state--action pairs. 

To allow for a similarity metric between discrete states, we model the state space as all possible combinations of $d$ binary variables, resulting in $N = 2^d$ possible states. Each of the $d$ outputs of the encoder defines the expectation of a Bernoulli random variable, with each variable sampled independently. The sampled states are used as input to the observation and transition networks, and as targets for the transition network.

The reconstruction and transition cost terms can now be used in
stochastic gradient descent on $\mathcal F$ in $\theta$, by estimating the gradient $\nabla_\theta \mathcal F$ with Monte Carlo samples from the variational posterior $q_\phi$.  To minimize $\mathcal F$ also in $\phi$, we need to perform backpropagation over discrete, stochastic variables (i.e. over $s^\mu_t$ sampled from $q_\phi$). There are several methods for doing this (see  Discussion). We use the reparameterization trick together with a relaxation of the Bernoulli distribution: the binary Con--crete (or Gumbel--Softmax) distribution (\citet{Maddison2016,jang2016categorical}). 


\subsection{The reparameterization trick and the Con--crete distribution}
Denoting the i\emph{th} dimension of state $s_t$ as $s_{t,i}$, we consider the i\emph{th} output of the encoder at time $t$ to correspond to $x_{t,i} = \logit(q_\phi(s_{t,i}=1|o_{t-k:t}))$. Following~\citet{Maddison2016}, we note that we can achieve a Bernoulli distribution by sampling according to $s_{t,i} = H(x_{t,i} + \mathit{L})$, where $H$ is the Heaviside step function and $\mathit{L}$ is a logistic random variable. In this form, the stochastic component $\mathit{L}$ is fully independent of $\phi$, and we can simply backpropagate through the deterministic nodes~\cite{kingma2013auto}. However, the derivative of $H$ is $0$ almost everywhere. To address this, the Bernoulli distribution can be relaxed into a continuous Con--crete (\emph{continuous} relaxation of \emph{discrete}) distribution~\cite{Maddison2016}. This corresponds to replacing the Heaviside non--linearity with a logistic non--linearity parameterized by the temperature $\lambda$:
\begin{equation}
\label{eq:concrete}
\hat{s}_{t,i} = \frac{1}{1 + \exp (-(x_{t,i} + \mathit{L})/\lambda)},
\end{equation}
with $\hat{s}_{t,i} \in [0,1]$. We use Con--crete samples from the encoder output for the input to both the reconstruction and transition networks and for the targets of the transition network, with temperatures taken from those suggested in~\cite{Maddison2016}: $\lambda_1 = 2/3$ for the posterior distribution and $\lambda_2 = 0.5$ for evaluating the transition log--probabilities. The Con--crete relaxation corresponds to replacing the discrete joint Bernoulli samples $s_t$ in the previous loss functions with their corresponding joint Con--crete samples $\hat{s}_t$. 
We train the network by sampling minibatches of observations and actions $(o^\mu_{t-k - 1:t}, a^\mu_t)$ from a replay memory~\cite{Riedmiller2005neural,mnih2015human} of transitions observed by the agent. 

\subsection{Learning a tabular transition model}

The model as described learns a joint Con--crete posterior distribution $\hat{q}_\phi(\hat{s}_{t}|o_{t-k:t})$. We can recover a discrete joint Bernoulli distribution $q_\phi(s_{t}|o_{t-k:t})$ by replacing the logistic non--linearity with a Heaviside non--linearity (i.e. as $\lambda \rightarrow 0$ in Eq. \ref{eq:concrete}).

For prioritized sweeping, we need to build a tabular model of the transition probabilities in the environment (i.e. $p(s_t| a_t, s_{t-1})$). We could consider extracting such a model from $p_{\theta_{\mathcal{T}}}(\hat{s}_t| a_t, \hat{s}_{t-1})$, the transition network used to train the encoder. However, this is problematic for several reasons. The transition network corresponds to Con--crete states, and is of a particularly simple form, where each dimension is sampled independently conditioned on the previous state and action. Moreover, the transition network is trained through stochastic gradient descent and therefore learns slowly; we want the agent to rapidly exploit new transition observations.

We therefore build a state transition table based purely on the encoder distribution $q_\phi(s_{t}|o_{t-k:t})$, by treating the most probable sequence of states under this distribution as observed data. Since each dimension of $s_t$ is independent conditioned on the observations, the mode $\bar{s}_t$ at time $t$ corresponds to a $d$--length binary string, where $\bar{s}_{t,i} = H(x_{t,i})$. Likewise, since states within an episode are assumed to be independent conditioned on the causal observation filter, the most probable state sequence for an episode is $\mathcal{S}^\mu = \{\bar{s}^\mu_0, \bar{s}^\mu_1\, \dots, \bar{s}^\mu_{T^\mu}\}$. We therefore record a transition between $\bar{s}_{t-1}$ and $\bar{s}_{t}$ under action $a_t$ for every step taken by the agent, and update the expected reward $\mathbb{E}[r|a_t, \bar{s}_{t-1}]$ in the table with the observed reward. Each binary string $\bar{s}$ is represented as a $d$--bit unsigned integer in memory. 

This process corresponds to empirically estimating the transition probabilities and rewards by counting, with counts that are revised during training. For instance, assume the agent encounters states $A$, $B$ and $C$ successively in the environment. We record transitions $A\rightarrow B$ and $B\rightarrow C$ in the table, and store the raw observations along with the corresponding state assignments $A$, $B$ and $C$ in the replay memory. If the observations associated with $B$ are later sampled from the replay memory and instead assigned to state $D$, we delete $A\rightarrow B$ and $B\rightarrow C$ from the table and add $A\rightarrow D$ and $D\rightarrow C$. Both the deletion and addition of transitions through training can change the $Q$-values.

\subsection{Using the model for reinforcement learning}
The $Q$-values in the table are updated continuously using the learned transition model $p(\bar{s}_t| a_t, \bar{s}_{t-1})$, expected rewards $\mathbb{E}[r|a_t, \bar{s}_{t-1}]$ and prioritized sweeping with small backups~\cite{van2013efficient}. Prioritized sweeping converges to the same solution as  value iteration, but can be much more computationally efficient by focusing updates on states where the $Q$-values change most significantly. 

Given an observation history $o_{t-k:t}$, the agent follows an $\epsilon$--greedy policy using the $Q$-values $Q(\bar{s}_t, a)$ in the lookup table for all possible actions $a$. For any pair $(\bar{s}_t, a)$ that has not yet been observed, we estimate the $Q$-value using an experience--weighted average over the nearest neighbours to $\bar{s}_t$ in Hamming distance (see Supplementary Materials for details). This Hamming neighbour estimate is parameter--less, and generally much faster than searching for nearest neighbours in continuous space.

\subsection{Implementation details}
The prioritized backups described by~\citet{van2013efficient} are performed serially with environment exploration. To decrease training time and improve performance, we performed backups independently, and in parallel, to environment exploration and training the deep network. 

We implemented state tabulation and prioritized sweeping as two separate processes (running on different CPU cores). The tabulation process acts in the environment and trains the neural networks by sampling the replay memory. The sweeping process maintains the transition table and continuously updates the Q-values using prioritized sweeping. 

To perform greedy actions, the tabulation process requests $Q$-values from the sweeping process. To update the transition table, the tabulation process sends transition updates (additions and deletions) to the sweeping process. Our implementation of the sweeping process performed ${\sim}6000$ backups/second, allowing the agent to rapidly propagate $Q$-value changes with little effect on the simulation time.

The pseudocode of \ac{VaST}, and of our implementation of prioritized sweeping, are in the Supplementary Material.

\begin{figure}[!ht]
    \includegraphics[width=\columnwidth]{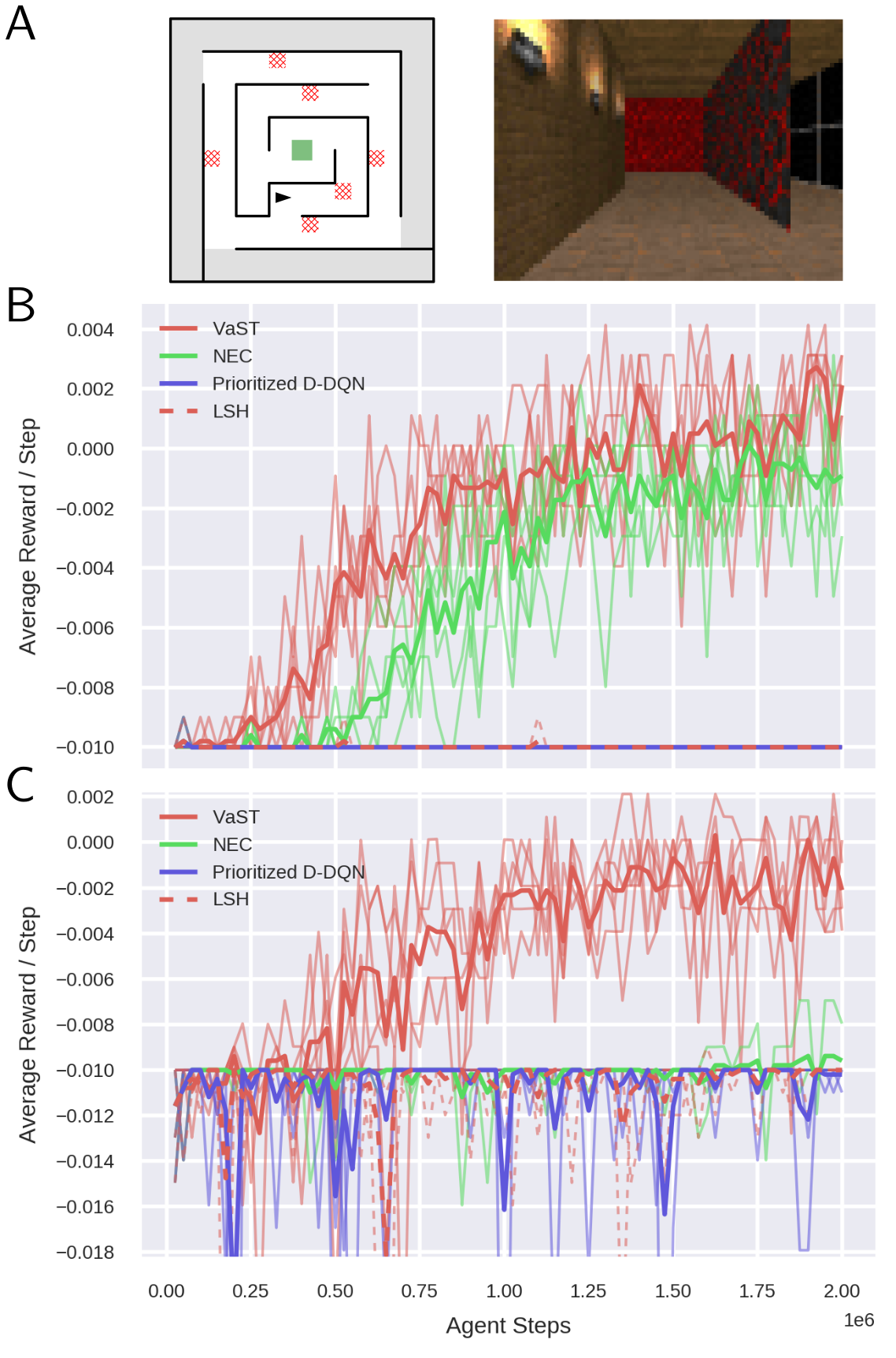}\vspace{-5pt}
    \caption{\textbf{\ac{VaST} learns quickly in complex mazes.} [\emph{A}]
        The agent started at a random position and orientation in the outer rim
        of the 3D maze (highlighted in grey), and received a reward of +1 on reaching the center of the maze (highlighted in green), with a step penalty of -0.01. Red hatched areas correspond to the hazard regions in the second version of the task, where the agent received a penalty of -1 with a probability of $25\%$. We used a different texture for each wall in the maze, ending at a corner. An example observation is shown for an agent positioned at the black arrow. [\emph{B}] Performance comparison between models for 5 individual runs with different random seeds (mean in bold). Rewards are very sparse ($\approx$ every 20\;000 steps with a random policy); with longer training we expect DQN to improve. [\emph{C}] Results for the second version of the task (including hazards).
    }\label{fig:hardmaze}
\end{figure}

\section{Results}
We evaluated the \ac{VaST} agent on a series of navigation tasks implemented in the \emph{VizDoom} environment (see \autoref{fig:hardmaze}A, \citet{Kempka2016ViZDoom}). Each input frame consists of a 3--channel $[60 \times 80]$ pixel image of the 3D environment, collected by the agent at a position $(x, y)$ and orientation $\theta$. The agent rarely observes the exact same frame from a previous episode ($0.05\%$ -- $0.3\%$ of the time in the mazes used here), making it ill--suited for a traditional tabular approach; yet the discovery of new transitions (particularly shortcuts) can have a significant effect on the global policy if leveraged by a model--based agent. We considered the relatively low--data regime (up to 2 million steps). Three actions were available to the agent: move forward, turn left and turn right; due to momentum in the game engine, these give rise to visually smooth trajectories. We also trained the agent on the Atari game Pong (\autoref{fig:pong}). For 3D navigation, we used only the current frame as input to the network, while we tested both 1-- and 4--frame inputs for Pong. 

We compared the performance of \ac{VaST} against two recently published sample--efficient model--free approaches: \ac{NEC}~\cite{Pritzel2017NEC} and Prioritized Double--DQN~\cite{Schaul2015prioritized}. We used the structure of the DQN network in~\cite{mnih2015human} for both \ac{NEC} and Prioritized D--DQN as well as the encoder of \ac{VaST} (excluding the output layers). Full hyperparameters are given in the Supplementary Material. 

We also compared against prioritized sweeping using \ac{LSH} with random projections~\citep{charikar2002similarity}, where each bit $\bar{s}_{t,i} = H(v_i \cdot o_{t})$, and each fixed projection vector $v_i$ had elements sampled from $\mathcal{N}(0,1)$ at the beginning of training. The environment model and $Q$-values were determined as with \ac{VaST}.

In the first task (\autoref{fig:hardmaze}), the agents were trained to reach a reward of +1 in the center of a complex maze, starting from a random position and orientation in the outer region. In a second version of the task, we added six ``hazard'' regions which gave a penalty of -1 with a probability of $25\%$ for each step. The agents were evaluated over a 1000--step test epoch, with $\epsilon=0.05$, every 25\;000 steps. \ac{VaST} slightly outperformed \ac{NEC} on the first version of the task and significantly outperformed all of the other models on the more difficult version (\autoref{fig:hardmaze}C).

\begin{figure}[!t]
    \includegraphics[width=\columnwidth]{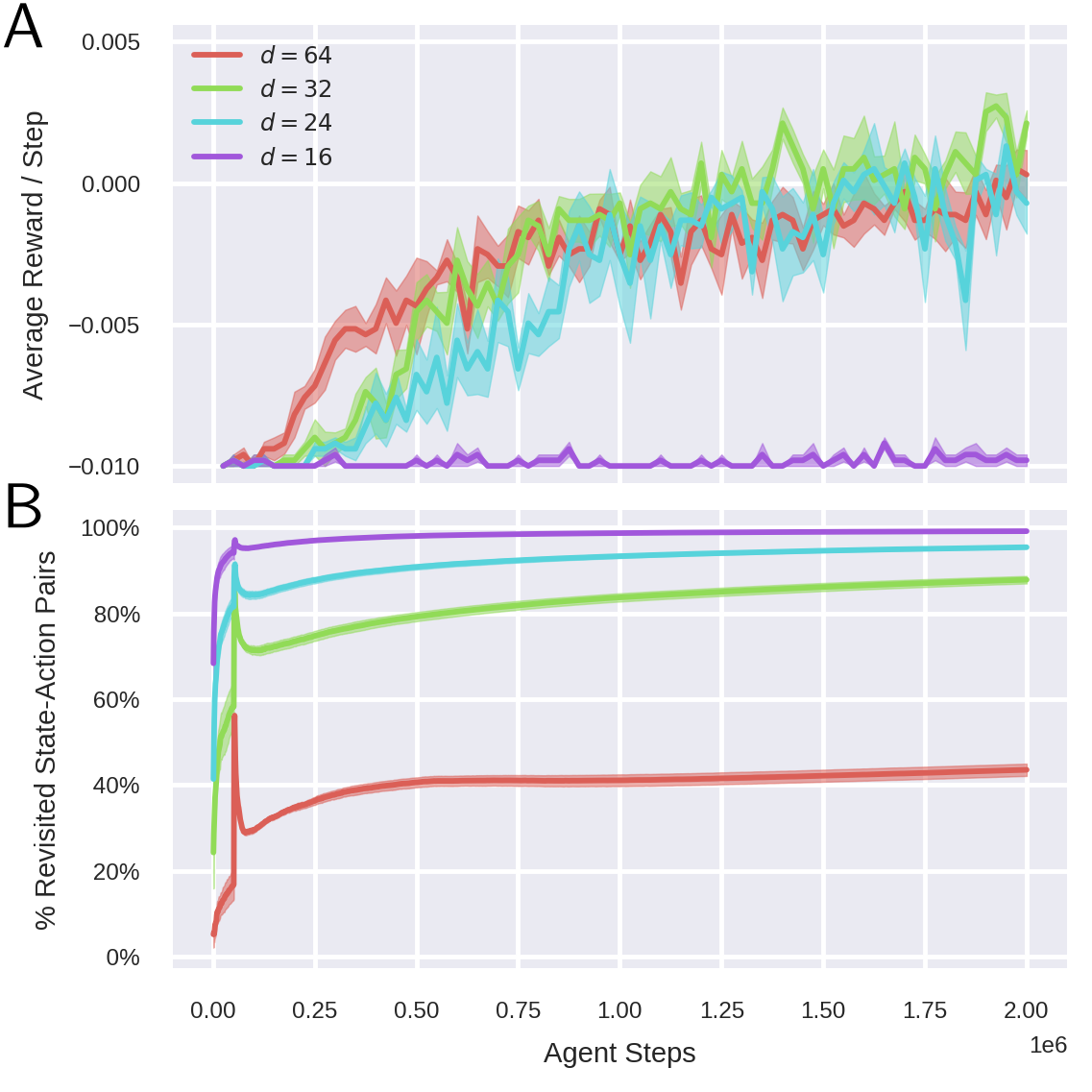}\vspace{-12pt}
    \caption{\textbf{Effect of latent dimensionality.} [\emph{A}] Average test reward $\pm$ SEM during training for $d=64$, $d=32$, $d=24$ and $d=16$ for the task in \autoref{fig:hardmaze}B (without hazards). [\emph{B}] Cumulative percentage of revisited state--action pairs during learning. The sharp transition at 50\;000 steps corresponds to the beginning of training the network.
    }\label{fig:compare}
\end{figure}

\subsection{Dimensionality of the latent representation}

We used $d=32$ latent dimensions for the \ac{VaST} agent in the navigation tasks, corresponding to a 32--bit representation of the environment. We examine the effect of $d$ in \autoref{fig:compare} and Supplementary Figure 1. High--dimensional representations ($d = 64$) tended to plateau at lower performance than representations with $d=32$, but also resulted in faster initial learning in the more complex maze. The agent frequently revisited state--action pairs even using the high dimensional representation (\autoref{fig:compare}B). In general, we found that we could achieve similar performance with a wide range of dimensionalities; smaller mazes could be learned with as few as 8--16 bits (Supplementary Figure 1). 



\begin{figure}[!t]
    \includegraphics[width=\columnwidth]{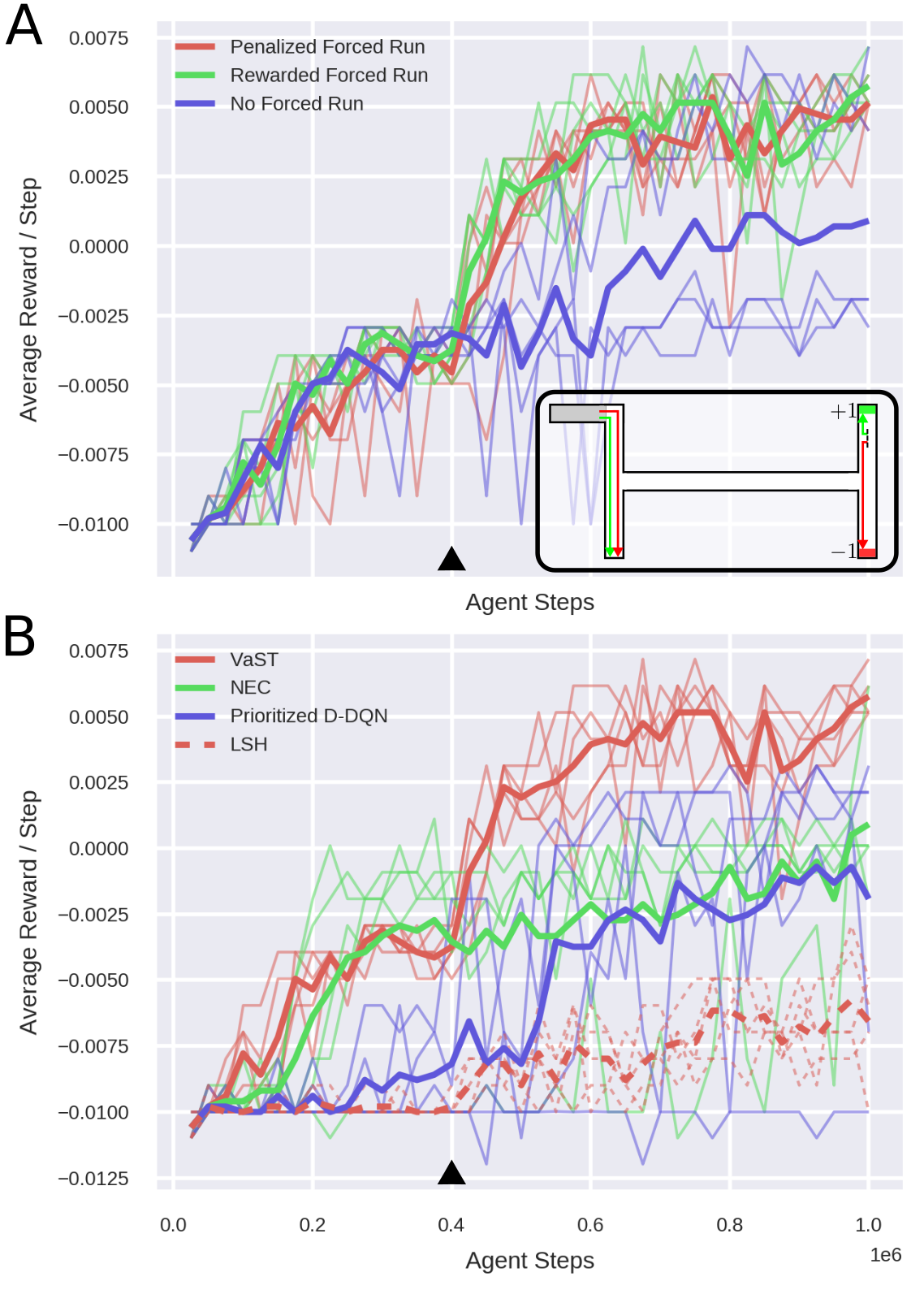}\vspace{-11pt}
    \caption{\textbf{\ac{VaST} allows for rapid policy changes in response to single experiences.} [\emph{A, Inset}] The agent learned to run from the starting area (grey) to a reward zone (green). After training, a new shortcut (teleporter) was introduced at the bottom of the left arm. The agent either observed no forced run, or a single forced run through the teleporter ending either in the rewarding (green) or the penalizing (red) terminal zone. The forced runs were 58 and 72 steps in length, respectively. [\emph{A}] The teleporter was introduced after 400\;000 steps (black triangle). The \ac{VaST} agent's performance is shown for the three conditions: no forced run, rewarded forced run and penalized forced run. [\emph{B}] Model performance comparison for rewarded forced runs. 
    }\label{fig:hmaze}
\end{figure}

\subsection{Sample efficiency}
We hypothesized that the \ac{VaST} agent would be particularly adept at rapidly modifying its policy in response to one new experience. To test this, we designed an experiment in a 3D H--maze (\autoref{fig:hmaze}) that requires the agent to leverage a single experience of a new shortcut. The agent learned to run towards a terminal reward zone (+1) while avoiding a dead end and a terminal penalty zone (-1), with a step penalty of -0.01. After 400\;000 steps of training (when the policy had nearly converged) we introduced a small change to the environment: running into the dead end would cause the agent to teleport to a position close to the reward zone, allowing it to reach the reward much faster. We informed the agent of the teleporter using a single forced run episode, in which the agent collected observations while running from the start box, through the teleporter, to either the reward zone or penalty zone under a fixed, predetermined policy. For the \ac{VaST} agent, this corresponds to a single experience indicating a new shortcut: the transition between the states before and  after the teleporter. After observing either the rewarded or penalized episode, performance rapidly improved as the agent adapted its policy to using the teleporter; in contrast, the agent discovered the teleporter on only 2/5 random seeds without the forced run. The agents switched to using the teleporter regularly approximately 20\;000 steps after the forced run, on average (about 160 episodes). The \ac{VaST} agent adapted to the teleporter more effectively than any of the other models (\autoref{fig:hmaze}B and Supplementary Figure 2).

\begin{figure}[!t]
    \includegraphics[width=\columnwidth]{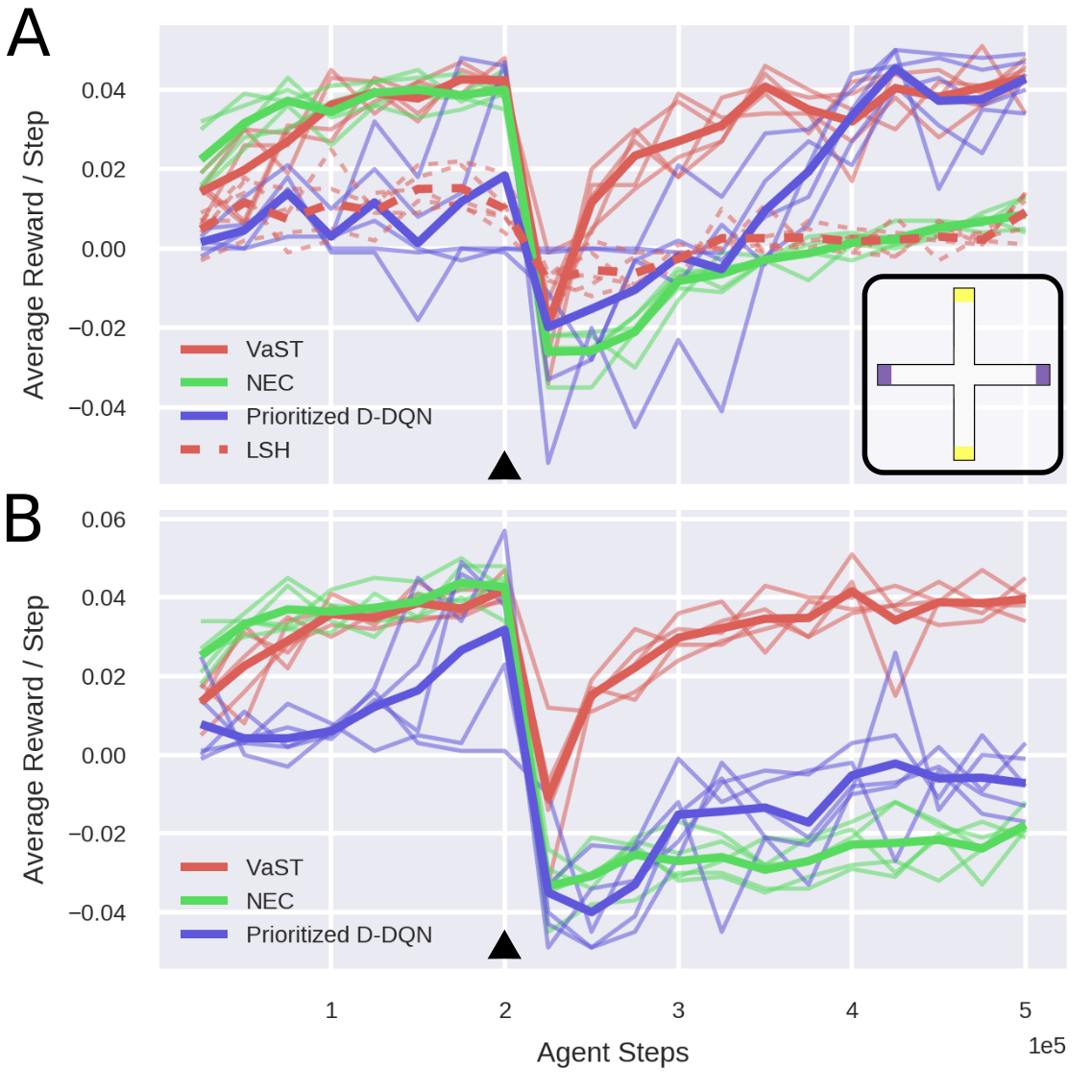}\vspace{-11pt}
    \caption{\textbf{\ac{VaST} can adapt to changing rewards.} [\emph{A, Inset}] The maze environment. Horizontal arms (purple) initially yielded a reward of +1 while vertical arms (yellow) yielded a penalty of -1. [\emph{A}] After training for 200\;000 steps (black triangle), the rewards and penalties in the maze were reversed. All agents used a replay memory size of $\mathcal N =$ 100\;000 transitions. [\emph{B}] The same task with a replay memory size of $\mathcal N =$ 500\;000.}
    \label{fig:plusmaze}
\end{figure}

\subsection{Transfer learning: non--stationary rewards}
\ac{VaST} keeps separate statistics on immediate rewards and transition probabilities in the environment. If the rewards were suddenly modified, we hypothesized that the existing transition model could allow the agent to rapidly adjust its policy (after collecting enough data to determine that the expected immediate rewards had changed).

We tested this in the maze shown in \autoref{fig:plusmaze}A (inset). Starting at a random position, the episode terminated at the end of any arm of the maze; the agent received a reward of +1 at the end of horizontal arms, and a penalty of -1 at the end of vertical arms. The reward positions were reversed after 200\;000 steps. We used two replay memory sizes ($\mathcal N=$ 100\;000 and $\mathcal N=$ 500\;000). Compared to \ac{NEC} and Prioritized D--DQN, \ac{VaST} both learned quickly in the initial phase and recovered quickly when the rewards were reversed. While both \ac{NEC} and Prioritized D--DQN adapted faster with a smaller replay memory, \ac{VaST} performed similarly in both conditions.

\begin{figure}[!t]
    \includegraphics[width=\columnwidth]{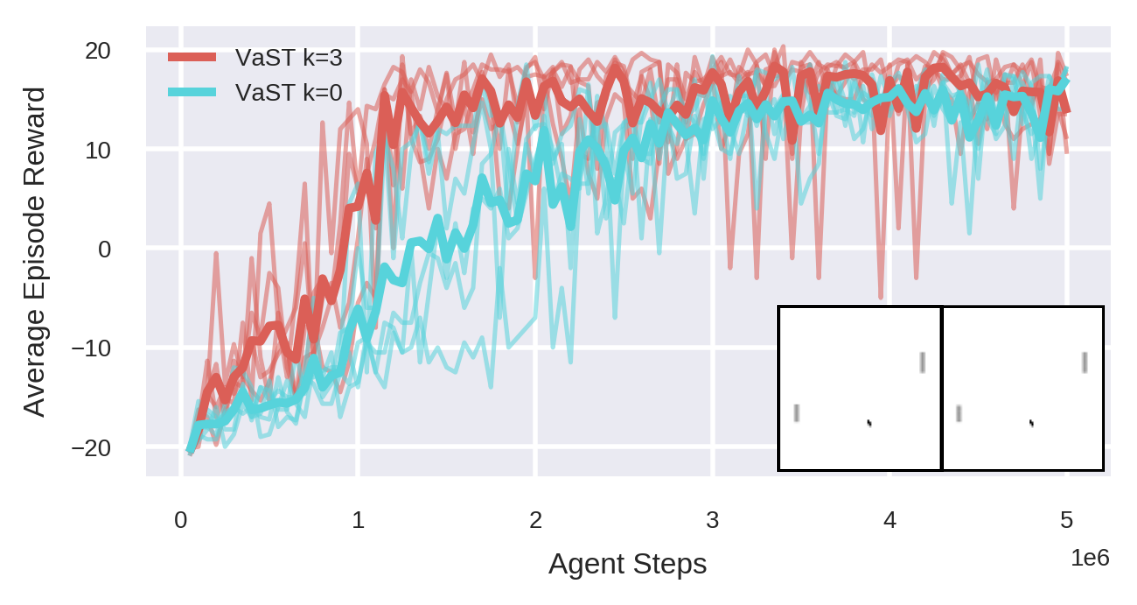}\vspace{-12pt}
    \caption{\textbf{Learning to play Pong.} Test epoch episode rewards for \ac{VaST} trained over 5 million steps. We tested performance with no frame history ($k=0$) and with 3 frames of history ($k=3$) as input to the encoder $q_\phi$. [\emph{Inset}] Actual observations $o_t$ (left) and reconstructed observations $\tilde{o}_t$ (right) for a trained agent.}
    \label{fig:pong}
\end{figure}

\subsection{Training on Atari: Pong}
In addition to 3D navigation, we trained the \ac{VaST} agent to play the Atari game Pong using the Arcade Learning Environment~\cite{bellemare13arcade}, with preprocessing steps taken from~\citet{mnih2013playing}. In Pong, a table tennis--like game played against the computer, the direction of the ball's movement is typically unclear given only the current frame as input. We therefore tried conditioning the posterior distribution $q_\phi$ on either the current frame ($k=0$) or the current frame along with the last 3 frames of input ($k=3$, following~\citet{mnih2013playing}). Using $k=3$, the performance converged significantly faster on average (\autoref{fig:pong}). While the reconstruction cost was the same for $k=0$ and $k=3$, the transition and entropy cost terms decreased with additional frame history (Supplementary Figure 3). 

\section{Related Work}
\paragraph{Model-based reinforcement learning}
Prioritized sweeping with small backups \cite{van2013efficient} is usually more efficient but similar to Dyna-Q \cite{Sutton2018}, where a model is learned and leveraged to update Q-values. Prioritized sweeping and Dyna-Q are background planning methods \cite{Sutton2018}, in that the action selection policy depends on Q-values that are updated in the background. In contrast, methods that rely on planning at decision time (like Monte Carlo Tree Search) estimate Q-values by expanding the decision tree from the current state up to a certain depth and using the values of the leaf nodes. Both background and decision time planning methods for model-based reinforcement learning are well studied in tabular environments \cite{Sutton2018}. Together with function approximation, usually used to deal with high--dimensional raw (pixel) input, many recent works have focused on planning at decision time.  \citet{Oh2017} and \citet{Farquhar2017}, extending the predictron \cite{Silver2017}, train both an encoder neural network and an action-dependent transition network on the abstract states used to run rollouts up to a certain depth. \citet{racaniere2017imagination} and \citet{Nagabandi2017} train a transition network on the observations directly. \citet{racaniere2017imagination} additionally train a rollout policy, the rollout encoding and an output policy that aggregates different rollouts and a model-free policy.  Planning at decision time is advantageous in situations like playing board games \cite{Silver2017A}, where the transition model is perfectly known, many states are visited only once and a full tabulation puts high demands on memory. Conversely, background planning has the advantage of little computational cost at decision time, almost no planning cost in well--explored stationary environments and efficient policy updates after minor environment changes. 

\paragraph{Successor representations for transfer learning}
The hybrid model--based/model--free approach of successor representations has recently been transferred from the tabular domain to deep function approximation~\cite{Dayan1993improving,Kulkarni2016deep}. 
Under this approach, the agent learns a model of the immediate reward from each state
and a model of the expected multi--step \emph{future occupancy} of each
successor state under the current policy.
As in a model--based approach, the immediate rewards can be updated independently of the environment dynamics.
However, the expected multi--step future occupancy is learned under a given policy, and the optimal policy will generally change with new rewards. 
The ability to generalize between tasks in an environment (as \ac{VaST} does in \autoref{fig:plusmaze}) then depends on the similarity between the existing and new policy. Recent work has proposed updating successor representations offline in a Dyna--like fashion using a transition model~\cite{Russek2017predictive,peng1993efficient}; we expect that prioritized sweeping with small backups could also be adapted to efficiently update tabular successor representations.

\paragraph{Navigation tasks}
Even though we demonstrate and evaluate our method mostly on navigation tasks,
VaST does not contain any inductive bias tailored to navigation problems. Using
auxiliary tasks \cite{Mirowski2016,jaderberg2016reinforcement}, we expect  further improvement in navigation.

\paragraph{State aggregation in reinforcement learning}
State aggregation has a long history in reinforcement learning 
\cite{li2006towards,Sutton2018}. To our knowledge, VaST is the first approach
that uses modern deep learning methods to learn useful and non-linear state
discretisation. In earlier versions of our model we tried discretising with \acp{VAE} as used by \cite{blundell2016episodic}, with mixed success. The state aggregator $q_\phi(s_t|o_{t-k:t})$ of VaST can be seen as a byproduct of training a hybrid ANN--HMM. Different methods to train ANN-HMMs have been studied 
\cite{Bengio1992,Tucker2017,Ng2016,Maddison2016}.  While none of these works
study the binary representation of the latent states used by \ac{VaST} for the
generalization of $Q$-values, we believe it is worthwhile to explore other training procedures and potentially draw inspiration from the ANN-HMM literature. 

%
%


\section{Discussion}

We found that the \ac{VaST} agent could rapidly transform its policy based on limited new information and generalize between tasks in the same environment. In stationary problems, \ac{VaST} performed better than competing models in complex 3D tasks where shortcut discovery played a significant role. Notably, \ac{VaST} performs \emph{latent learning}; it builds a model of the structure of the environment even when not experiencing rewards~\cite{Tolman1930maze}.

We also trained \ac{VaST} to play the Atari game Pong. In general, we had less initial success training the agent on other Atari games. We suspect that many Atari games resemble deterministic tree Markov Decision Processes, where each state has exactly one predecessor state. In these tasks, prioritized sweeping conveys no benefit beyond MFEC~\cite{brea2017}. In contrast, intrinsically continuous tasks like 3D navigation can be well--characterized by a non--treelike tabular representation (e.g. by using a discretisation of ($x$,$y$,$\theta$), where $\theta$ denotes the agent's orientation).

\ac{VaST} differs from many deep reinforcement learning models in that the neural network is entirely reward--agnostic, where training corresponds to an unsupervised learning task. Many other possible architectures exist for the unsupervised tabulator; for instance, a Score Function Estimator such as NVIL~\cite{Mnih2014,Mnih2016A,Tucker2017} could be used in place of the Con--crete relaxation for discrete stochastic sampling. In addition, while we chose here to show the strengths of a purely model--based approach, one could also consider alternative models that use value information for tabulation, resulting in hybrid model--based/model--free architectures. 

The past decade has seen considerable efforts towards using deep networks to adapt tabular \ac{RL} techniques to high--dimensional and continuous environments. Here, we show how the opposite approach -- using deep networks to instead transform the environment into a tabular one -- can enable the use of powerful model--based techniques.


\section*{Acknowledgements}

We thank Vasiliki Liakoni and Marco Lehmann for their invaluable suggestions, feedback and corrections on the manuscript. This work was supported by the Swiss National Science Foundation (grant agreement no. 200020\_165538).

\bibliographystyle{icml2018}
\bibliography{writeup}

\end{document}


\maketitle
\section{VaST pseudocode}
\begin{algorithm}
	\caption{Variational State Tabulation.
				\label{alg:VaST}}
	Initialize replay memory $\mathcal{M}$ with capacity $\mathcal{N}$\\
	Initialize sweeping table process $\mathcal{B}$ with transition add queue $\mathcal{Q}^+$ and delete queue $\mathcal{Q}^-$
   \begin{algorithmic}[1]
	\FOR{each episode}
		\STATE Set $t \leftarrow 0$
		\STATE Get initial observations $o_0$\\
		\STATE Process initial state $\bar{s}_0 \leftarrow \argmax_{s} q_\phi(s|o_{0})$\\
		\STATE Store memory $(o_{0},\bar{s}_{0})$ in $\mathcal{M}$
		\WHILE{not terminal}
			\STATE Set $t \leftarrow t + 1$
			\STATE Take action $a_{t}$ with $\epsilon$-greedy strategy based on $\tilde{Q}(s_{t-1},a)$ from $\mathcal{B}$
			\STATE Receive $r_t$, $o_t$
			\STATE Process new state $\bar{s}_t \leftarrow \argmax_{s} q_\phi(s|o_{t-k:t})$
			\STATE Store memory $(o_{t},\bar{s}_{t},a_{t},r_{t})$ in $\mathcal{M}$
			\STATE Put transition $(\bar{s}_{t-1},a_{t},r_{t},\bar{s}_{t})$ on $\mathcal{Q}^+$
			\IF{training step}
				\STATE Set gradient list $\mathcal{G} \leftarrow \{\}$
				\FOR{sample in minibatch}
					\STATE Get $(o_{j-k - 1:j}, a_j)$ from random episode and step $j$ in $\mathcal{M}$
					\STATE Process $q_\phi(s_{j-1}|o_{j-k-1:j-1})$, $q_\phi(s_j|o_{j-k:j})$ with encoder
					\STATE Sample $\hat{s}_{j-1}$, $\hat{s}_{j} \sim \hat{q}_\phi$ with temperature $\lambda$
		  			\STATE Process $p_\theta(o_j|\hat{s}_j)$, $p_\theta(\hat{s}_j|a_j,\hat{s}_{j-1})$ with decoder and transition network
		  			\STATE Append $\nabla_{\theta, \phi} \mathcal{F}(\theta, \phi; o_{j-k - 1:j})$ to $\mathcal{G}$
					\FOR{$i$ in \{$j-1$, $j$\}}
						\STATE Process $\bar{s}^{new}_i \leftarrow \argmax_{s} q_\phi(s|o_{i-k:i})$
						\STATE Get ($\bar{s}_{i-1}$, $a_{i}$, $r_i$, $\bar{s}_i$, $a_{i+1}$, $r_{i+1}$, $\bar{s}_{i+1}$) from $\mathcal{M}$
						\IF{$\bar{s}_{i} \neq \bar{s}^{new}_{i}$}
							\STATE Put $(\bar{s}_{i-1},a_{i},r_{i},\bar{s}_{i})$,  $(\bar{s}_{i},a_{i+1},r_{i+1},\bar{s}_{i+1})$ on $\mathcal{Q}^-$
							\STATE Put $(\bar{s}_{i-1},a_{i},r_{i},\bar{s}^{new}_{i})$,  $(\bar{s}^{new}_{i},a_{i+1},r_{i+1},\bar{s}_{i+1})$ on $\mathcal{Q}^+$
							\STATE Update $\bar{s}_{i} \leftarrow \bar{s}^{new}_{i}$ in $\mathcal{M}$
						\ENDIF
					\ENDFOR
				\ENDFOR
				\STATE Perform a gradient descent step according to $\mathcal{G}$ with given optimizer
			\ENDIF
	  	\ENDWHILE
	\ENDFOR
   \end{algorithmic}
\end{algorithm}

\section{Details to prioritized sweeping algorithm}
We follow the ``Prioritized Sweeping with reversed full backups'' algorithm~\citep{van2013efficient} with some adjustments: a subroutine is added for transition deletions, and priority sweeps are performed continuously except when new transition updates are received. The $Q$-values of unobserved state--action pairs are never used, so we simply initialize them to $0$.  Finally, we kept a model of the expected immediate rewards $\mathbb{E}[r|s,a]$ explicitly, although this is not necessary and was not used in any of the experiments presented; we omit it here for clarity.

In the algorithm, discretised states $\bar{s}$ are simplified to $s$.

\begin{algorithm}
    \caption{Prioritized Sweeping Process.}\label{alg:ps}
	Initialize $V(s) = U(s) = 0$ for all s \\
	Initialize $Q(s,a) = 0$ for all s, a \\
	Initialize $N_{sa}, N^{s'}_{sa} = 0$ for all $s$, $a$, $s'$ \\
	Initialize priority queue $\mathcal{P}$ with minimum priority cutoff $p_{min}$\\
	Initialize add queue $\mathcal{Q}^+$ and delete queue $\mathcal{Q}^-$
   \begin{algorithmic}[1]
   \WHILE{True}
		\WHILE{$\mathcal{Q}^+$, $\mathcal{Q}^-$ empty}
			\STATE Remove top state $s'$ from $\mathcal{P}$
			\STATE $\Delta U \leftarrow V(s') - U(s')$
			\STATE $U(s') \leftarrow V(s')$
			\FOR{$\mathbf{all}$ $(s,a)$ pairs with $N^{s'}_{sa} > 0$}
				\STATE $Q(s,a) \leftarrow Q(s,a) + \gamma N^{s'}_{sa}/N_{sa}\cdot \Delta U$
				\STATE $V(s) \leftarrow \max_b\{Q(s, b)|N_{sb}>0\}$
				\STATE add/update $s$ in $\mathcal{P}$ with priority $| U(s) - V(s) |$ if $| U(s) - V(s) | > p_{min}$
			\ENDFOR
		\ENDWHILE
		\FOR{$(s,a,r,s')$ in $\mathcal{Q}^+$}
			\STATE $N_{sa} \leftarrow N_{sa} + 1$; $N^{s'}_{sa} \leftarrow N^{s'}_{sa} + 1$
			\STATE $Q(s,a) \leftarrow [Q(s,a)(N_{sa} - 1) + r + \gamma U(s')]/N_{sa}$
			\STATE $V(s) \leftarrow \max_b\{Q(s, b)|N_{sb}>0\}$
			\STATE add/update $s$ in $\mathcal{P}$ with priority $| U(s) - V(s) |$ if $| U(s) - V(s) | > p_{min}$
		\ENDFOR
		\FOR{$(s,a,r,s')$ in $\mathcal{Q}^-$}
			\STATE $N_{sa} \leftarrow N_{sa} - 1$; $N^{s'}_{sa} \leftarrow N^{s'}_{sa} - 1$
			\IF{$N_{sa} > 0$}
				\STATE $Q(s,a) \leftarrow [Q(s,a)(N_{sa} + 1) - (r + \gamma U(s'))]/N_{sa}$
			\ELSE
				\STATE $Q(s,a) \leftarrow 0$
			\ENDIF
			\IF{$\sum_bN_{sb} > 0$}
				\STATE $V(s) \leftarrow \max_b\{Q(s, b)|N_{sb}>0\}$
			\ELSE
				\STATE $V(s) \leftarrow 0$
			\ENDIF
			\STATE add/update $s$ in $\mathcal{P}$ with priority $| U(s) - V(s) |$ if $| U(s) - V(s) | > p_{min}$
		\ENDFOR
	\ENDWHILE
   \end{algorithmic}
\end{algorithm}
\newpage

\section{Details to $Q$-value estimation}

Here, we simplify the discretised states $\bar{s}$ to $s$ for clarity. We denote $\mathcal{S}$ as the set of all states corresponding to $d$--length binary strings, $\tilde{Q}(s, a)$ as the $Q$-value estimate used for action selection, and $Q(s, a)$ as the $Q$-value for a state--action pair in the lookup table as determined by prioritized sweeping (which is only used if $(s,a)$ has been observed at least once).

In order to calculate $\tilde{Q}(s_t, a)$ for a particular state--action pair, we first determine the Hamming distance $m$ to the nearest neighbour(s) $s \in \mathcal{S}$ for which the action $a$ has already been observed, i.e.
\begin{align}
\label{eq:hamdist}
m &= \min_{s \in \mathcal{S}}\{D(s_t, s)| N_{sa} > 0\},
\end{align}
where $D(s_t,s)$ is the Hamming distance between $s_t$ and $s$ and $N_{sa}$ denotes the number of times that action $a$ has been taken from state $s$. We then define the set $\mathcal{S}_{tm}$ of all $m$--nearest neighbours to state $s_t$, 
\begin{align}
\label{eq:lookupset}
\mathcal{S}_{tm} &= \{s \in \mathcal{S} | D(s_t, s) = m\},
\end{align}
and the $Q$-value estimate used for action selection is then given by
\begin{equation}
\label{eq:lookup}
\tilde{Q}(s_t,a) := \dfrac{\sum_{s \in \mathcal{S}_{tm}} N_{sa}Q(s,a)}{\sum_{s \in \mathcal{S}_{tm}}N_{sa}}.
\end{equation}

If $(s_t, a)$ has already been observed, then $m=0$, $\mathcal{S}_{tm} = \{s_t\}$ and $\tilde{Q}(s_t,a) = Q(s_t,a)$. If $m=1$, $\tilde{Q}(s_t,a)$ corresponds to an experience--weighted average over all states $s$ with a Hamming distance of 1 from $s_t$, $m=2$ to the average over neighbours with a Hamming distance of 2 etc.

$\tilde{Q}(s_t,a)$ can be seen as the $Q$-value of an abstract aggregate state $s_{tm}$ consisting of the $m$--nearest neighbours to $s_t$. To show this, we introduce the index set of past experiences $\mathcal E_{sa} = \{(\tau, \mu) | s_\tau^\mu = s, a_\tau^\mu = a\}$ that contains all the time indices $\tau$ for all episodes $\mu$ where action $a$ was chosen in state $s$ (taking into account all reassignments as described in section 2.3 of the main text and in \autoref{alg:VaST}). With the above definition of $N_{sa}$ we see that $N_{sa} = |\mathcal E_{sa}|$, i.e. there are $N_{sa}$ elements in the set $\mathcal E_{sa}$. With this and the update mechanism of prioritized sweeping (\autoref{alg:ps}) we can write
\begin{align}
\label{eq:psstate}
Q(s,a) = \frac{1}{N_{sa}}\sum_{\tau, \mu \in \mathcal E_{sa}}r_\tau^\mu + \gamma\frac{1}{N_{sa}}\sum_{\tau, \mu \in \mathcal E_{sa}}V(s_{\tau+1}^\mu),
\end{align}
where $V(s) = \max_b\{Q(s, b)|N_{sb}>0\}$. Substituting this into~\autoref{eq:lookup}, we obtain
\begin{equation}
\label{eq:1aggstate}
\tilde{Q}(s_t,a) = \dfrac{\sum_{s \in \mathcal{S}_{tm}} \Big[\sum_{\tau, \mu \in \mathcal E_{sa}}r^\mu_\tau + \gamma\sum_{\tau, \mu \in \mathcal E_{sa}}V(s_{\tau+1}^\mu)\Big]}{\sum_{s \in \mathcal{S}_{tm}}N_{sa}}.
\end{equation}
We now consider an aggregate state $s_{tm}$ by treating all states $s \in \mathcal{S}_{tm}$ as equivalent, i.e. $\mathcal E_{s_{tm}a} = \{(\tau, \mu) | s_\tau^\mu \in \mathcal{S}_{tm}, a_\tau^\mu = a\}$. With this definition we get $\sum_{s \in \mathcal{S}_{tm}} \sum_{\tau, \mu \in \mathcal E_{sa}} =  \sum_{\tau, \mu \in \mathcal E_{s_{tm}a}}$ and we obtain
\begin{align}
\label{eq:2aggstate}
\tilde{Q}(s_t,a) &= \dfrac{\Big[\sum_{\tau, \mu \in \mathcal E_{s_{tm}a}}r^\mu_\tau + \gamma\sum_{\tau, \mu \in \mathcal E_{s_{tm}a}}V(s_{\tau+1}^\mu)\Big]}{N_{s_{tm}a}} \\
\nonumber &= Q(s_{tm},a), 
\end{align}
where we used \autoref{eq:psstate} to obtain the second equality.

\newpage
\section{Extended latent dimensionality analysis}
\begin{figure}[!h]
\includegraphics[width=\columnwidth]{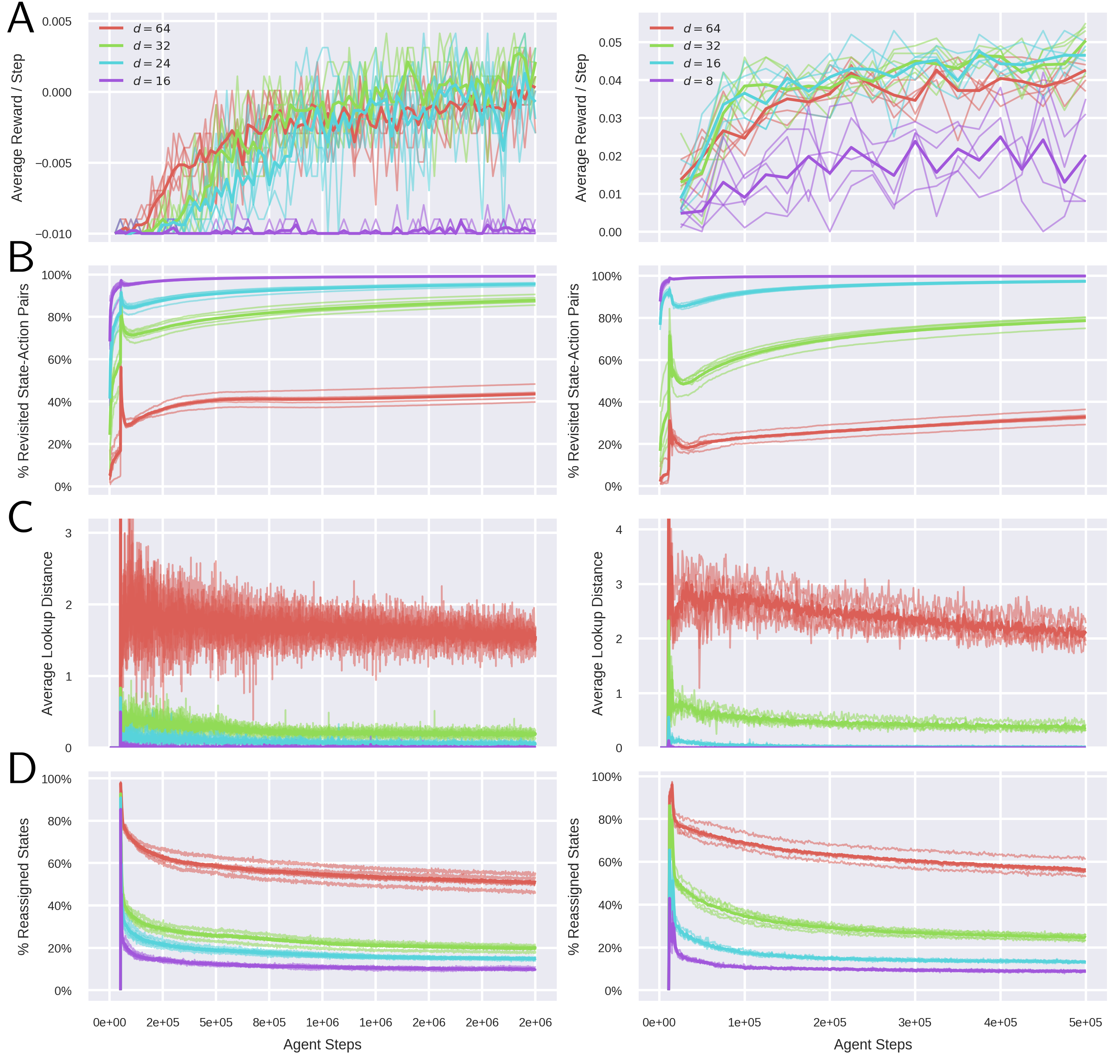}
    \caption{Effect of latent dimensionality in a large maze (left column, Figure 3B in main text) and a small maze (right column, Figure 6 in main text). [\emph{A}] Average reward. [\emph{B}] Cumulative percentage of revisited state--action pairs over the course of training. The sharp transition at 50\;000 steps corresponds to the beginning of training. [\emph{C}] The average lookup distance $m$ as a function of time. [\emph{D}] The average percentage of observations from a minibatch that were reassigned to a different state during training.
    }\label{fig:compare}
\end{figure}
\newpage

\section{Extended sample efficiency results}
\begin{figure}[!h]
\centering
\includegraphics[width=0.9\columnwidth]{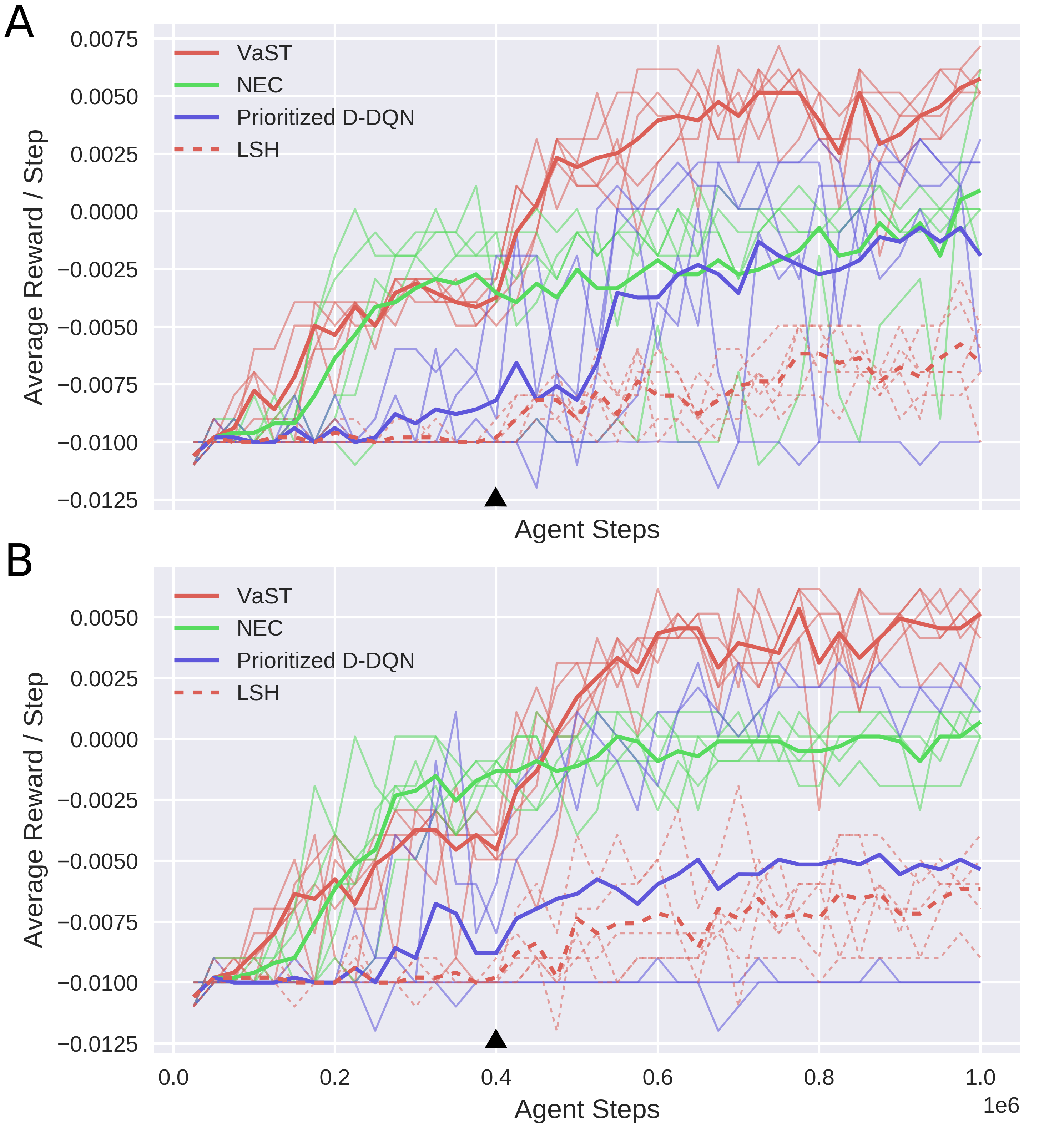}
    \caption{Performance comparison between models for [\emph{A}] rewarded forced runs (identical to Figure 5B in main text) and [\emph{B}] penalized forced runs. Black arrows indicate addition of teleporter and forced runs.}
\end{figure}
\newpage

\section{Effect of training on frame histories}
\begin{figure}[!h]
\includegraphics[width=\columnwidth]{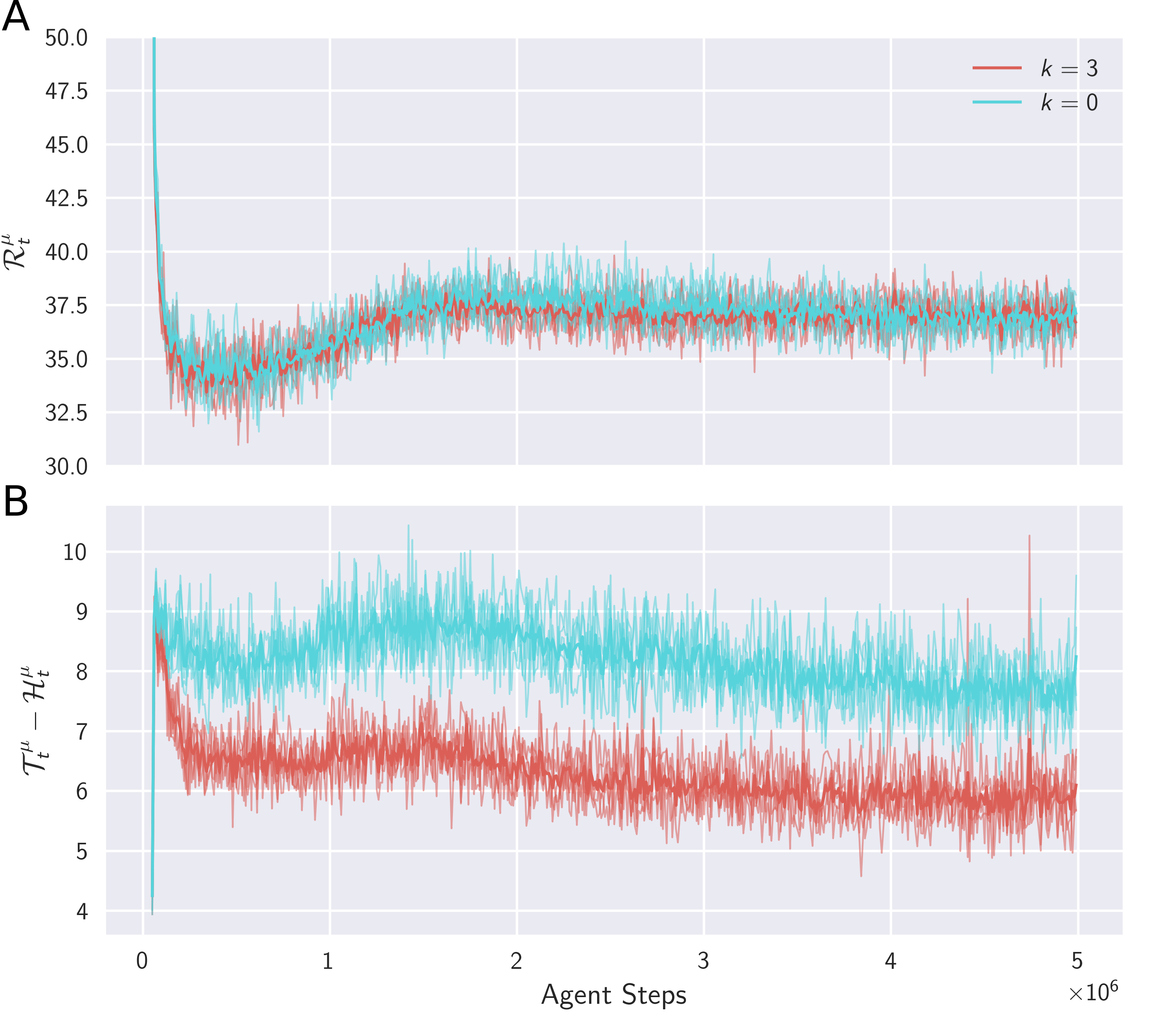}
	\caption{The free energy cost function over the course of training on Pong, broken into [\emph{A}] the reconstruction terms and [\emph{B}] the transition and entropy terms, conditioning on three additional past frames of observations ($k=3$) and no additional frames ($k=0$). Training with past frames as input resulted in faster learning on Pong (main text, Figure 7). As shown here, training on past frames conveys no added benefit in reconstructing the current frame, but instead decreases the additional cost terms.}
\end{figure}

\section{Hyperparameters}

\subsection{3D Navigation}

For the three network--based models, hyperparameters were chosen based on a coarse parameter search in two mazes (Figure 3 excluding the hazards and Figure 5 excluding the teleporter), using the previously published hyperparameters as a starting point for the baselines~\citep{Pritzel2017NEC,Schaul2015prioritized,mnih2015human}. In all mazes except the smaller Plus--Maze, the agents explored randomly for 50\;000 steps to initialize the replay memory before training; $\epsilon$ was then annealed from $1$ to $0.1$ over 200\;000 steps. In the Plus--Maze, the agents explored randomly for 10\;000 steps and $\epsilon$ was annealed over 40\;000 steps. We used $\epsilon = 0.05$ for evaluation during test epochs, which lasted for 1000 steps. In all tasks we used a discount factor of $0.99$. 

The encoder of VaST and the networks for NEC and Prioritized D--DQN all shared the same architecture, as published in~\citep{mnih2015human}, with ReLU activations. For all three networks, we used the Adam optimizer~\citep{kingma2014adam} with $\beta_1 = 0.9$, $\beta_2 =0.999$, and $\epsilon = 1\mathrm{e}{-8}$, and trained on every $4$th step. Unless otherwise stated, we used a replay memory size of $\mathcal{N} =$ 500\;000 transitions. 

\subsubsection{VaST}

We used a latent dimensionality of $d=32$ unless otherwise stated. For training, we used a minibatch size of $128$ and a learning rate of $2 \times 1\mathrm{e}{-4}$. For sweeping, we used $p_{min} = 5 \times 1\mathrm{e}{-5}$. For the Con--crete relaxation, we used the temperatures suggested by~\cite{Maddison2016}: $\lambda_1 = 2/3$ for sampling from the posterior and evaluating the posterior log--probability and $\lambda_2 = 0.5$ for evaluating the transition and initial state log--probabilities.

For the decoder architecture, we used a fully--connected layer with $256$ units, followed by 4 deconvolutional layers with $4 \times 4$ filters and stride $2$, and intermediate channel depths of $64$, $64$ and $32$ respectively. We used an MLP with 3 hidden layers (with $512$, $256$ and $512$ units respectively) for each action in the transition network.

\subsubsection{NEC}

We used a latent embedding of size $64$, $n_s = 50$ for the n--step $Q$-value backups, and $\alpha = 0.1$ for the tabular learning rate. We performed a $50$ approximate nearest--neighbour lookup using the ANNoy library (pypi.python.org/pypi/annoy) on Differentiable Neural Dictionaries of size 500\;000 for each action. For training, we used a minibatch size of $32$ and a learning rate of $5 \times 1\mathrm{e}{-5}$. 

\subsubsection{Prioritized D--DQN}

We used the rank--based version of Prioritized DQN with $\alpha=0.7$ and $\beta=0.5$ (annealed to 1 over the course of training). We used a minibatch size of 32 and a learning rate of $1\mathrm{e}{-4}$ and updated the target network every 2000 steps.

\subsubsection{LSH}

The LSH--based algorithm does not use a neural network or replay memory, since the embedding is based on fixed random projections. We achieved the best results with $d=64$ for the latent dimensionality. For prioritized sweeping, we used $p_{min} = 5 \times 1\mathrm{e}{-5}$.

\subsection{Atari: Pong}
We used a latent dimensionality of $d=64$, a replay memory size of $\mathcal{N} =$ 1\;000\;000 transitions, and annealed $\epsilon$ over 1\;000\;000 steps. All other hyperparameters were the same as for navigation.

\bibliographystyle{plainnat}
\bibliography{writeup}